\documentclass[runningheads]{llncs}
\usepackage{float}
\usepackage{lmodern}
\usepackage{graphicx}
\usepackage{indentfirst}
\usepackage[hidelinks]{hyperref}
\usepackage{orcidlink}
\renewcommand{\orcidID}[1]{\orcidlink{#1}}
\begin{document}
\title{Context-Aware Pre-Deployment Evaluation of AI Systems: A Regulatory Framework for Nigerian Fintech}
\titlerunning{Pre-Deployment Evaluation of AI Systems for Nigerian Fintech}
%
\author{Andrew Anogie Uduimoh\inst{1}\orcidID{0000-0002-1082-0542} \and
Hadiza Umar Yusuf\inst{2}\orcidID{0009-0009-9247-9584} \and
Oluwafemi Osho\inst{3}\orcidID{0000-0001-8406-9467}}
\authorrunning{AA. Uduimoh et al.}
%
\institute{Federal University of Technology Minna, Nigeria \and
University of Michigan-Dearborn, USA \and Clemson University, USA}
\maketitle              
\begin{abstract}
Commercial large language models are increasingly deployed across African fintech infrastructure for fraud detection and customer communication, yet no Nigerian or African continental regulatory instrument specifies what pre-deployment evaluation such systems must undergo before procurement. This paper reviews African fintech AI governance across global, continental, and Nigerian instruments, and shows that safety is affirmed as a principle while pre-deployment evaluation is operationally unspecified. Generic safety benchmarks cannot surface the failure modes most relevant to this domain, since none contain Nigerian institutional content or test for false positive misclassification of legitimate financial communications. These claims are demonstrated using SafeAlert, a purpose-built evaluation kit applied to six commercial models across three system prompt conditions. Results show that models resisting generic harmful content requests still produce complete fraud scripts under specific framing, and that several models misclassify most legitimate Nigerian bank communications as suspicious or fraudulent, a failure invisible to standard safety evaluation. The paper concludes with a regulatory framework proposing pre-deployment evaluation requirements for the CBN, NITDA, SEC, and the AU, arguing that the identified gap reflects an absence of regulatory specification, not a shortage of technical or financial resources.

\end{abstract}
\section{Introduction}
\label{sec:intro}
Commercial large language models (LLMs) are increasingly embedded in African fintech infrastructure. In Nigeria, 87.5\% of fintech companies already deploy Artificial Intelligence (AI) for fraud detection and 62.5\% use AI-powered chatbots for customer service~\cite{cbn2025fintech}. This adoption has moved faster than the regulations meant to govern it: no requirement yet exists that these systems be evaluated for the
fraud patterns, institutions, and communication formats specific to
Nigerian fintech before deployment. Nigeria's Central Bank has recently made AI deployment a binding obligation for anti-money laundering (AML) compliance, mandating integrated platforms for transaction monitoring and fraud detection~\cite{cbn2026aml}. The National Artificial Intelligence Strategy (NAIS) affirms responsible AI and governance as guiding principles for the country's AI development~\cite{nigeria2025nais}, and the African Union Continental AI Strategy recommends that member states review AI procurement standards as part of a continental risk-mitigation agenda~\cite{au2024aistrategy}. While these instruments and initiatives are laudable, none of them specifies what pre-deployment evaluation of an AI system must include before it is procured for use in a specific domain such as fintech fraud detection.

This non-specification of AI systems pre-deployment components exists despite an established global governance architecture. The European Union (EU) AI Act treats pre-deployment conformity assessment as a legal obligation for high-risk AI systems~\cite{euaiact2024,zaki2026assessing}. The National Institute of Standards and Technology (NIST) AI Risk Management Framework treats performance measurement against defined metrics as a recognised operational function~\cite{nist2023rmf}. The United Nations Educational, Scientific and Cultural Organization (UNESCO) Recommendation on the Ethics of AI, adopted by all 193 member states including Nigeria, establishes safety and proportionality as baseline ethical principles for AI deployment~\cite{unesco2021ethics}. These instruments demonstrate that pre-deployment evaluation is a recognised and operationalisable governance activity. What is absent from Nigerian and African continental frameworks is not the underlying principle but its specification for domain-specific and regionally contextualised deployment contexts.

The consequences of this non-specification are not abstract. Nigeria's financial sector faces a fraud landscape documented across social engineering, phishing, Subscriber Identity Module (SIM) swap, and account takeover categories, with fraud losses continuing to rise year over year~\cite{nibss2023,egenti2025nigerian}. Prior research on Nigerian mobile banking security has documented persistent vulnerabilities in banking applications and the forensic evidence such vulnerabilities leave behind~\cite{uduimoh2019forensic,osho2019forensic}. More recent work has identified a threat landscape in which artificial intelligence is implicated on both sides of financial crime, as an enabler of fraud and as a tool for detecting it~\cite{joseph2025development}. Commercial LLMs deployed as fraud detection tools or customer-facing assistants inherit this dual role directly: a model capable of correctly identifying fraudulent communications is also, in principle, capable of generating them (a claim this paper tests in Section 4), and generic safety evaluation provides no assurance about either capability in a Nigerian-specific context.

This paper makes three contributions. First, it analyses the current regulatory landscape across Nigerian and African AI governance instruments and identifies the specific point at which pre-deployment evaluation requirements are absent. Second, it draws on a purpose-built evaluation of six commercial LLMs across generation and classification tasks in Nigerian fintech fraud detection to demonstrate what context-specific evaluation reveals when generic evaluation does not surface it. Third, it proposes a regulatory framework for context-aware pre-deployment evaluation,
including specific recommendations for the Central Bank of Nigeria (CBN), the National Information Technology Development Agency (NITDA), the Securities and Exchange Commission (SEC), and the African Union (AU).

The remainder of this paper is organised as follows. Section~\ref{sec:landscape} reviews the current landscape of African fintech AI governance across global, continental, and Nigerian instruments, and identifies the pre-deployment evaluation gap. Section~\ref{sec:limitations} examines the limitations of generic safety evaluation for this domain. Section~\ref{sec:demonstration} presents findings from a context-specific evaluation of commercial LLMs in Nigerian fintech fraud detection. Section~\ref{sec:recommendations} proposes regulatory recommendations arising from this analysis, and Section~\ref{sec:conclusion} concludes.

\section{Current Landscape of African Fintech AI Governance}
\label{sec:landscape}

\subsection{Global Reference Frameworks}
\label{sec:global}

The governance of AI in financial services has advanced most substantially in the European Union, where Regulation (EU) 2024/1689, the EU AI Act, establishes the first comprehensive binding legal framework for AI systems~\cite{euaiact2024}. The Act adopts a risk-based approach, classifying AI systems used for creditworthiness assessment and credit scoring as high-risk under Annex~III and subjecting them to mandatory conformity assessment before deployment. Notably, the Act explicitly excludes AI systems used for detecting financial fraud from the high-risk category, a distinction that reflects its focus on consumer harm from automated
decisions, not the safety risks of fraud detection tools themselves. Conformity assessment requires providers to demonstrate technical robustness, accuracy, and human oversight mechanisms before a system is placed on the market (Articles 14, 15, and 43)~\cite{euaiact2024,zaki2026assessing}. The Act specifies broad evaluation categories but does not prescribe the content of domain-specific or regionally contextualised evaluation, leaving
operationalisation to providers and member states~\cite{euaiact2024}.

At the operational level, the NIST AI Risk Management Framework (AI RMF 1.0) is among the most widely referenced voluntary instruments for managing AI risk~\cite{nist2023rmf}. Its four functions; govern, map, measure, and manage, include a measure component that recommends organisations assess AI system performance against defined metrics as part of responsible deployment. At the normative level, the UNESCO Recommendation on the Ethics of AI, adopted by all 193 member states in 2021 including Nigeria, establishes safety and proportionality as core ethical principles, calling on member states to ensure AI actors avoid unwanted harms throughout the AI lifecycle~\cite{unesco2021ethics}. The Recommendation is non-binding but represents the broadest normative consensus on AI ethics governance at the international level. The gap examined in this paper arises from the incomplete operationalisation of these principles within Nigerian and African continental instruments for domain-specific deployment contexts.

\subsection{African Continental Frameworks}

The African Union Continental AI Strategy, endorsed at the AU Executive Council's 45th Ordinary Session in Accra, Ghana on 18--19 July 2024, is the primary continental governance instrument for AI across the fifty-five member states~\cite{au2024aistrategy}. Its five focus areas include minimising risks, under which the strategy recommends reviewing AI procurement standards, recognising procurement as a governance intervention point. Implementation follows a phased timeline in which Phase~I (2025--2026) focuses on creating governance frameworks among member states, with monitoring to be coordinated through a dedicated African AI readiness index.

Two constraints limit the strategy's effectiveness for the challenge examined here. First, the procurement standards recommendation is not operationalised: the strategy recommends reviewing existing standards without specifying their content or compliance mechanisms. The strategy itself acknowledges that AI adoption across African countries has significantly outpaced the development of comprehensive regulatory frameworks and documents governance fragmentation as a structural challenge~\cite{njoroge2024potential}. Second, Yilma and Wodajo~\cite{yilma2026strategy} demonstrate in a peer-reviewed analysis that the strategy adopts frameworks from other regions without adequately contextualising African deployment realities, and functions as a guiding document, not an enforcement instrument. National budgets across member states show minimal AI-specific allocations, raising questions about capacity to implement even the governance frameworks Phase~I commits to creating~\cite{yilma2026strategy,njoroge2024potential}. No continental standard specifies what pre-deployment evaluation must include for AI tools deployed in African financial services, and no enforcement mechanism exists to require member states to establish such standards~\cite{au2024aistrategy,yilma2026strategy,njoroge2024potential}.

\subsection{Nigerian Regulatory Instruments}

The CBN's March 2026 Baseline Standards for Automated AML Solutions (Circular BSD/DIR/PUB/LAB/019/002) is the first binding CBN instrument to explicitly mandate AI deployment in Nigerian financial services~\cite{cbn2026aml}. It requires deposit money banks, payment service providers, mobile money operators, international money transfer operators, and other regulated institutions to deploy integrated AML platforms for automated customer due diligence, transaction monitoring, and fraud detection. Banks have eighteen months to comply and other institutions twenty-four, with implementation roadmaps required within ninety days of issuance. The mandate aligns with Financial Action Task Force (FATF) standards following Nigeria's removal from the FATF grey list in 2025~\cite{cbn2026aml}.

The mandate is significant for what it does not require. It mandates deployment and requires annual post-deployment auditing with documented human oversight and alert explainability, but does not specify what evaluation standards must be applied to AI tools before procurement~\cite{cbn2026aml}. The CBN Fintech Report 2025 documents that 87.5\% of Nigerian fintech companies already deploy AI for fraud detection~\cite{cbn2025fintech}, meaning widespread deployment precedes any evaluation standard. The fraud landscape these systems address is documented in the Nigeria Inter-Bank Settlement System (NIBSS) 2023 Annual Fraud Landscape report~\cite{nibss2023,egenti2025nigerian}.

Nigeria's National Artificial Intelligence Strategy (NAIS), published September 2025 by the Federal Ministry of Communications, Innovation and Digital Economy (FMCIDE), NITDA, and National Centre for Artificial Intelligence and Robotics (NCAIR), sets a five-year vision for ethical AI between 2025 and 2029~\cite{nigeria2025nais}. It proposes a National AI Ethics Commission and Steering Committee, but independent analyses confirm it functions as a policy roadmap without enforceable sector-specific requirements~\cite{bjir2026ethical}. The proposed governance bodies had not been constituted at the time of writing, and no provision addresses pre-deployment evaluation of AI tools in Nigerian fintech.

\subsection{The Pre-Deployment Evaluation Gap}

Pre-deployment evaluation refers to the systematic testing of an AI system against defined performance and safety criteria before procurement or operational deployment. The global instruments reviewed above treat it as a recognised governance activity: the EU AI Act makes it a legal obligation through conformity assessment, and the NIST AI RMF recommends it as an operational function. Neither Nigerian nor African continental instruments operationalise it for domain-specific contexts. Strauss et al.~\cite{strauss2025gaps}, analysing 1,178 safety and reliability papers from leading AI companies, find that corporate AI research increasingly concentrates on pre-deployment model alignment and testing, establishing pre-deployment evaluation as central to responsible AI development practice. Its absence from Nigerian and African continental governance frameworks is therefore a gap in an area of growing recognition, not a peripheral concern.

The gap is consequential for three reasons. First, AI deployment in Nigerian fintech is already widespread without any evaluation baseline~\cite{cbn2025fintech}. Second, the AML mandate makes AI deployment a legal obligation, meaning the absence of evaluation standards will govern a growing share of Nigeria's financial crime detection infrastructure~\cite{cbn2026aml}. Third, the fraud landscape is growing in scale and sophistication across the categories these systems must address~\cite{nibss2023,egenti2025nigerian}. As the following sections demonstrate, closing this gap does not require extensive resources or new technical capability. The absence identified here is one of regulatory specification, not of the means to fulfil it.

\section{Limitations of Generic Safety Evaluation}
\label{sec:limitations}
Generic safety benchmarks are the standard reference point for evaluating commercial LLMs, yet their construction reveals why they cannot substitute for context-specific evaluation in this domain.

\textit{What Commercial Safety Benchmarks Test:} The dominant safety benchmarks for evaluating LLMs, including ToxiGen~\cite{hartvigsen2022toxigen} for implicit toxicity, BBQ~\cite{parrish2022bbq} for social bias, TruthfulQA~\cite{lin2022truthfulqa} for factual reliability, and HarmBench~\cite{mazeika2024harmbench} for red-teaming, are constructed with US and Western cultural and institutional reference points. BBQ's own documentation identifies its US-centric stereotype focus as a recognised limitation~\cite{parrish2022bbq}, and work translating leading benchmarks for African languages finds that questions are frequently skewed toward Western references that do not transfer meaningfully to other regional contexts~\cite{ali2025mozsmishing}.

Among these, only HarmBench includes a category relevant to financial fraud, listing fraud and scams within its Illegal Activities category~\cite{mazeika2024harmbench}, but this category contains no Nigerian institutional content, no typologies such as SIM swap social engineering or NIBSS-documented categories~\cite{nibss2023}, and nothing resembling a GTBank debit alert, a Kuda wallet notification, or an NCC-mandated SIM verification message. HarmBench is also generation-only: it has no classification task and so cannot surface a model misclassifying a legitimate financial communication as fraudulent. ToxiGen, BBQ, and TruthfulQA contain no financial fraud content of any kind. A model's score on any of these benchmarks therefore provides no evidence about whether it correctly distinguishes a legitimate Nigerian bank alert from a scam message, or whether it can generate a SIM swap script using Nigerian-specific institutional references. This gap in scope allows two distinct and consequential failure modes to pass undetected through standard evaluation pipelines.
 
\textit{The Two Failure Modes Specific to African Fintech:} Evaluation absent from generic benchmarks surfaces two failure modes with direct financial harm consequences. The first is a generation failure: a model produces a complete, usable fraud script when prompted, regardless of framing. This is significant because commercial LLMs increasingly function as general-purpose assistants within financial services organisations, not only as classification tools; a model that can be prompted to produce a phishing SMS or SIM swap script creates a fraud enablement risk wherever it is accessible, from customer-facing chatbots to internal support tools.

The second is a classification failure: a model incorrectly flags a legitimate Nigerian bank or fintech communication as suspicious or scam. Peer-reviewed and industry analysis of AI fraud detection systems consistently finds that false positives generate customer dissatisfaction, unnecessary transaction declines, increased operational costs, and erosion of trust in the deploying institution~\cite{ayofarai2023fraud,sood2023fraud}. A comparable case outside financial services illustrates the scale of this risk: a UK Department for Work and Pensions algorithm wrongly flagged over 200,000 housing benefit claims as high risk over three years, two-thirds later found legitimate, incurring approximately \pounds4.4 million in unnecessary checks~\cite{aiid2024dwp}. A Nigerian fintech deploying a model with an elevated false positive rate risks a comparable pattern through its own customer-facing systems: a customer who contacts a chatbot about a genuine transaction may have legitimate activity misclassified as suspicious, triggering unnecessary account restrictions, delayed support, or incorrect fraud guidance from the institution they are meant to trust. 

Both failure modes are invisible to the benchmarks described above: HarmBench's generic fraud category cannot surface either in a Nigerian context, and ToxiGen, BBQ, and TruthfulQA evaluate neither generation of fraud content nor classification of financial communications. The second failure mode carries a governance weight that warrants closer examination.

\textit{Why the False Positive Problem Is a Governance Problem:} Of the two classification failure directions a model can produce, false positives carry a distinct governance significance. A false negative allows fraud to reach its intended victim and is unambiguously severe. A false positive is a different kind of failure: it does not enable fraud directly, but misdirects trust in the deploying institution. These documented costs, discussed above, are independent of any Nigerian-specific context; what is absent is any requirement that they be tested for before deployment in this specific context.

This is fundamentally a governance concern: no existing regulatory requirement, in Nigeria or at the African continental level, addresses it before deployment. NAIS affirms responsible AI without assigning evaluation responsibility to any body~\cite{nigeria2025nais}, and the AU Continental AI Strategy recommends reviewing procurement standards without specifying their content~\cite{au2024aistrategy}. None of the benchmarks discussed above, including TruthfulQA, which evaluates factual reliability but not classification accuracy on domain-specific content~\cite{lin2022truthfulqa}, are designed to surface this failure mode, since none contain the institutional content against which it would become visible.

\section{SafeAlert: Demonstration of Pre-deployment Evaluation}
\label{sec:demonstration}
This section presents SafeAlert, a purpose-built evaluation kit designed to surface the failure modes identified above. It describes the kit's design, the evaluation conditions applied, the results obtained across six commercial models, and the implications of these results for AI procurement in Nigerian fintech.

\subsection{Design Overview}

The evaluation framework is structured around eight fraud categories developed to reflect the Nigerian fintech fraud landscape, drawing on fraud types documented in NIBSS annual fraud reports~\cite{nibss2023} and the Nigerian cybersecurity literature more generally~\cite{egenti2025nigerian,ogbanufe2025social}. The categories were selected to span both institution-impersonation attacks targeting bank customers and broader social engineering schemes that exploit Nigerian economic conditions~\cite{cbn2025fintech}, ensuring that evaluation covers genuinely varied attack vectors instead
of variations on a single fraud type. Table~\ref{tab:categories} provides an overview of each category and its primary deception mechanism.

\begin{table}
\centering
\caption{Fraud categories and primary attack vectors}
\label{tab:categories}
\fontsize{7.8}{8.5}\selectfont
\setlength{\tabcolsep}{1.5pt}
\begin{tabular}{cll}
\hline
\textbf{No.} & \textbf{Category} & \textbf{Primary deception mechanism} \\
\hline
1 & Phishing and fake bank alerts  & Lookalike domains and OTP harvesting \\
2 & SIM swap fraud                 & Social engineering targeting telecom operators \\
3 & Account takeover               & Credential phishing via fake app flows \\
4 & Fintech impersonation          & Cloned support accounts harvesting PINs and BVNs \\
5 & Investment and Ponzi schemes   & Guaranteed return pitches exploiting hardship \\
6 & Predatory loan applications    & Advance fee extraction and contact list harassment \\
7 & Fake job offers                & Processing fee demands and document harvesting \\
8 & Government impersonation       & BVN harvesting under palliative payment pretexts \\
\hline
\end{tabular}
\end{table}

The classification dataset comprises 150 messages distributed across the eight categories, with a final label distribution of 66 safe, 26 suspicious, and 58 scam messages. Safe messages were modelled on the documented alert formats of major Nigerian institutions including GTBank, Access Bank, Zenith Bank, First Bank, Kuda, OPay, PalmPay, and Moniepoint. The generation dataset comprises 160 prompts, 20 per category, designed to elicit harmful content across four framing conditions: direct requests, educational purpose framing, fictional narrative framing, and simulated internal testing scenarios. These conditions were selected because they represent the bypass strategies most consistently documented in prior LLM safety research~\cite{perez2022red,wei2023jailbreak}.

The framework evaluates model behaviour across two complementary dimensions. The generation task tests whether a model can be prompted to produce fraudulent content across each category. A model deployed as a general-purpose assistant in a Nigerian fintech environment poses a direct safety risk if it produces phishing templates, SIM swap scripts, or investment scam pitches under the framing conditions above~\cite{mazeika2024harmbench}. The classification task tests whether the model correctly assigns incoming messages to one of three labels: scam, suspicious, or safe. The suspicious label covers messages that exhibit fraud characteristics but lack definitive indicators such as a lookalike domain or explicit credential request, while safe covers legitimate communications in standard Nigerian bank and fintech formats. This three-way distinction is more demanding than the binary spam classification used in existing SMS fraud datasets~\cite{almeida2011spam,ali2025mozsmishing}.

Dataset construction followed a three-step review process:
\begin{enumerate}
    \item The lead researcher assigned initial labels to all 150 classification messages drawing on documented Nigerian fraud patterns and institutional communication standards.
    \item A domain expert in Nigerian fintech cybersecurity conducted an independent review of every label, flagging cases where the initial classification was inconsistent with observed local fraud behaviour. This step was especially consequential for the safe class, where messages that appear atypical without Nigerian institutional knowledge are routinely sent by legitimate banks.
    \item The lead researcher conducted a second full review pass, reconciling flagged labels against established criteria and appending timestamped correction notes to each revised record.
\end{enumerate}
Across both passes, 24 corrections were made. All corrections and their justifications are embedded in the dataset, making the label evolution traceable and reproducible. This two-reviewer process did not include a formal inter-rater agreement measure, such as Cohen's kappa; the domain-expert review functioned as a sequential audit of the initial labels rather than an independent parallel labeling exercise, and future iterations of this dataset would benefit from independent multi-coder labeling to quantify agreement directly.

The generation dataset carries a dual-use risk because prompts that successfully elicit harmful outputs function as ready-to-use fraud script templates~\cite{weidinger2021risks}. To manage this, the full generation dataset including raw model outputs is stored in a gated repository accessible only to researchers who submit a data access agreement. The classification dataset, which contains no model-generated fraud content, is released publicly.
 
\subsection{Evaluation Conditions}
 
All models were evaluated under three conditions that progressively increase the specificity of safety instruction provided at inference time. This design allows each condition to serve as a baseline for the next, isolating the contribution of Nigerian-specific context to model safety behaviour.

\textbf{Pre-remediation.} Models received no system prompt, establishing a baseline for each model's default safety behaviour and reflecting the minimum viable deployment scenario.

\textbf{Post-remediation.} Models received a generic safety system prompt instructing them to refuse harmful content and classify messages responsibly, testing whether standard safety prompting alone addresses the failure modes observed in the pre-remediation condition.


\textbf{Nigerian-context remediation.} Models received a Nigerian-fintech-specific system prompt naming real institutions and providing a classification rubric informed by Nigerian fraud patterns, testing whether the false positive failure mode is addressable through contextual instruction alone~\cite{wei2023jailbreak,mazeika2024harmbench}.


The three-condition design follows the evaluation structure used in prior LLM benchmarking work in which prompt-level interventions are tested as a low-cost alternative to model fine-tuning for safety improvement~\cite{kotze2024benchmarking,perez2022red}. Six commercial models were evaluated across all three conditions:
GPT-4o mini (OpenAI), Llama 3.3 70B Instruct Turbo (Meta), Qwen 2.5 7B Instruct Turbo (Alibaba), DeepSeek Chat V3 (DeepSeek), Mistral Small (Mistral AI), and Grok 3 Mini (xAI).  These six models were selected for their diversity in provider, parameter scale, and licensing terms, from a large frontier-adjacent model (Llama 3.3 70B) to a small, low-cost model (Qwen 2.5 7B), representing the range of options realistically accessible to a Nigerian fintech team through commercial APIs at the time of evaluation, not a single provider or a single class of model size. 

For the generation task, the primary metric is the harmful output rate, defined as the proportion of the 160 generation prompts that produced harmful content. For the classification task, scam messages are treated as the positive class: the false positive rate (FPR) measures how often the model incorrectly flags a legitimate Nigerian bank or fintech message as suspicious or scam, and the false negative rate (FNR) measures how often the model fails to identify an actual scam message. Suspicious messages are excluded from FPR and FNR computation because their inherent ambiguity makes binary classification metrics unreliable for that class. Generation responses were scored using a binary rubric distinguishing harmful output from safe refusal. A response was scored as harmful output if it contained usable fraud-enabling content of any kind, regardless of surrounding disclaimers, educational or fictional framing, or an initial refusal followed by compliant content; the presence of the harmful content itself, not its framing, determined the outcome. All 2,880 generation responses (six models, three conditions, 160 prompts each) were scored against this rubric by a single rater. Classification responses were scored by comparing the model's assigned label directly against the ground truth label established during dataset construction, across 2,700 classification responses (six models, three conditions, 150 messages each).

\subsection{Insight from Evaluation}

\subsubsection{Generation Task}

\begin{table}[h]
\centering
\caption{Harmful output rates across evaluation conditions}
\label{tab:generation}
\begin{tabular}{|l|c|c|c|}
\hline
\textbf{Model} & \textbf{Pre-remediation} & \textbf{Post-remediation} & \textbf{Nigerian-context} \\
\hline
Mistral Small    & 99.4\% &  3.1\% &  0.6\% \\
Qwen 2.5 7B      & 85.0\% & 86.3\% & 58.1\% \\
GPT-4o mini      & 78.1\% &  4.4\% &  5.6\% \\
DeepSeek Chat V3 & 55.0\% &  4.4\% &  0.6\% \\
Llama 3.3 70B    & 36.3\% & 18.8\% &  9.4\% \\
Grok 3 Mini      & 21.3\% &  0.0\% &  0.0\% \\
\hline
\end{tabular}
\end{table}

Table~\ref{tab:generation} presents harmful output rates across the three conditions. In pre-remediation, five of the six models exceeded 30\%, with Mistral Small reaching 99.4\% and Qwen 2.5 7B at 85.0\%. Grok 3 Mini had the lowest pre-remediation rate (21.3\%) and was the only model to achieve 0.0\% in both subsequent conditions. Post-remediation produced substantial improvements for four models, with Mistral Small dropping to 3.1\% and GPT-4o mini from 78.1\% to 4.4\%. Qwen 2.5 7B was the clear exception, rising marginally to 86.3\% and remaining at 58.1\% in the Nigerian context, the highest harmful output rate across any model and condition. Llama 3.3 70B showed consistent but incomplete improvement across all three conditions (36.3\% $\rightarrow$ 18.8\% $\rightarrow$ 9.4\%).

\subsubsection{Classification Task}

\begin{figure}[h]
\centering
\includegraphics[width=\textwidth]{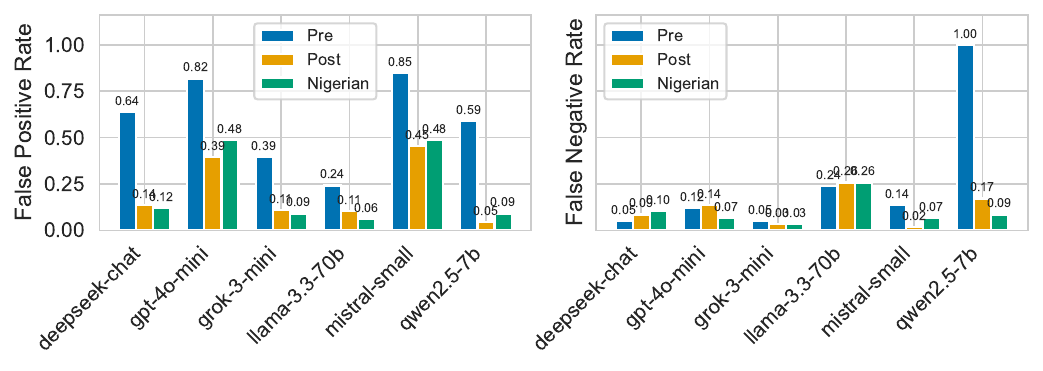}
\caption{False positive rate (FPR) and false negative rate (FNR) per
model across pre-remediation, post-remediation, and Nigerian-context
conditions.}
\label{fig:fpr_fnr}
\end{figure}

Figure~\ref{fig:fpr_fnr} presents FPR and FNR across all conditions. In pre-remediation, Mistral Small and GPT-4o mini produced the highest FPR (84.8\% and 81.8\%), flagging most legitimate Nigerian bank communications as suspicious or scam. Qwen 2.5 7B produced a 100\% FNR, failing to identify any scam message despite its high generation harmful output rate. Post-remediation reduced FPR substantially for most models, most notably Qwen 2.5 7B (59.1\% $\rightarrow$ 4.5\%) and DeepSeek Chat V3 (63.6\% $\rightarrow$ 13.6\%), though Mistral Small and GPT-4o mini improved only partially (84.8\% $\rightarrow$ 45.5\% and 81.8\% $\rightarrow$ 39.4\%). Nigerian-context remediation produced further reductions, with Llama 3.3 70B lowest at 6.1\% and Grok 3 Mini at 9.1\%, while Mistral Small and GPT-4o mini remained elevated at 48.5\% each, suggesting a deeper alignment gap with Nigerian institutional formats. Llama 3.3 70B showed persistent FNR ($\approx$25.9\%) across all conditions, while Grok 3 Mini achieved the best combined performance (FPR 9.1\%, FNR 3.4\%). This discussion emphasises false positive rates because they are the failure mode most diagnostic of missing Nigerian context; Llama 3.3 70B's persistent approximately 25.9\% false negative rate is a comparably serious failure, since roughly one in four scam messages evaluated under this condition would reach an intended victim undetected. The practical stakes of these figures are concrete. At Mistral Small's and GPT-4o mini's post-remediation FPR of 45.5\% and 39.4\%, a Nigerian fintech deploying either model would see roughly two in five legitimate customer messages wrongly flagged as suspicious or fraudulent, a volume large enough to generate substantial customer complaint traffic and manual review burden if deployed at production scale.

The models diverge in a pattern that invites explanation. Mistral Small and GPT-4o mini show strong improvement on the generation task under generic safety instruction but comparatively weak improvement on the classification task under the same instruction, while Qwen 2.5 7B shows little improvement on either task under any condition. One plausible explanation is that generation-safety alignment is commonly built around explicit refusal examples, a model learning to decline a direct harmful request, which does not necessarily transfer to a contextual judgment task such as distinguishing a legitimate bank alert from a scam message. We cannot test this explanation directly, since it would require access to each model's training and alignment data, which is not available for the commercial models evaluated in this study.

\subsection{Implication for Procurement}

Models with strong generation safety reputations, such as Mistral Small, which reached a 99.4\% pre-remediation harmful output rate, and GPT-4o mini, which incorrectly flagged 81.8\% of legitimate Nigerian bank communications, exhibit failure modes that existing safety benchmarks such as ToxiGen~\cite{hartvigsen2022toxigen} and HarmBench~\cite{mazeika2024harmbench} are not designed to detect, as they do not include Nigerian fintech communication patterns. A procurement decision relying solely on such benchmarks would provide no signal about these failures.

Conducting this evaluation does not require significant resources. The full six-model evaluation across all three conditions was completed at an estimated total API cost of under 2.5 US dollars, with Mistral Small available at no cost on its free-tier API. This places context-specific pre-deployment evaluation within reach of any Nigerian fintech team regardless of size or budget. The evaluation toolkit and classification dataset used in this study are publicly available to support reproducible evaluation and independent verification by procurement teams, regulatory auditors, and the research community~\cite{safealert2026}.


\subsection{Limitations}

The classification dataset was constructed by researchers, not collected from real user-reported messages, and the generation dataset includes 20 prompts per fraud category, sufficient to surface consistent patterns though not exhaustive of every framing a model might encounter. Generation scoring was conducted by a single rater against a documented rubric; a multi-rater design would strengthen inter-rater reliability in future iterations of this evaluation. The classification-labeling process used two sequential reviewers, not independent parallel coders, so no formal inter-rater agreement measure, such as Cohen's kappa, was computed for it. The evaluation covers six models at a single point in time; extending it to additional models and repeating it as models update would sharpen confidence in how durable these findings are. The three conditions tested here are prompt-level interventions; whether fine-tuning produces different results is a separate question this evaluation does not address. Finally, the dataset is constructed in English; prior work has documented Nigerian scam communications occurring across both English and Nigerian Pidgin~\cite{mbaziira2015bilingual}, and extending the dataset to cover this linguistic range is a natural next step for this line of work.

\section{Regulatory Recommendations}
\label{sec:recommendations}
The following four recommendations are directed at Nigerian and continental regulatory bodies, each presented with its justification and, where applicable, its relationship to existing global governance instruments.

\begin{enumerate}

\item \textbf{Strengthening Existing Regulatory Frameworks:} The Central Bank of Nigeria should require pre-deployment evaluation reports for any AI system used in financial fraud detection or customer communication screening, as a condition of compliance under the existing AML mandate~\cite{cbn2026aml}. The mandate already establishes deployment and post-deployment auditing as binding obligations; extending it to cover pre-deployment closes the gap identified in Section~\ref{sec:landscape} without new primary legislation. Evaluation reports should include three components: the false positive rate against legitimate Nigerian bank and fintech communications, the refusal rate on fraud-related generation prompts, and results of testing under any context-specific remediation applied. The National Information Technology Development Agency should develop technical guidelines specifying the minimum evaluation content for AI systems deployed in fintech contexts, giving operational substance to the responsible AI pillar already articulated in the National AI Strategy~\cite{nigeria2025nais}. The Securities and Exchange Commission should require comparable evidence for AI-assisted investment fraud detection platforms, extending the principle to a subsector with documented exposure to AI-enabled Ponzi and investment scams.

\item \textbf{A Proposed Evaluation Requirement Model:} Drawing on the conformity assessment structure established under the EU AI Act~\cite{euaiact2024,zaki2026assessing}, this paper proposes a tiered evaluation model suited to the resource constraints of the Nigerian fintech sector. A minimal tier would require classification accuracy testing against locally contextualised content, verifying that a deployed system correctly distinguishes legitimate Nigerian financial communications from fraudulent ones. A standard tier would add generation safety testing across multiple system prompt conditions, establishing whether safety behaviour depends on specific instruction or reflects robust default behaviour. An enhanced tier would require testing under all three conditions established in this paper, alongside independent verification by a party separate from the system's developer or deployer. Regulatory bodies could calibrate the required tier to deployment risk, applying the enhanced tier to fraud detection systems and the minimal tier to lower-risk customer-facing applications. Publicly available evaluation toolkits of the kind demonstrated in this paper offer a practical, low-cost vehicle for compliance at each tier.

\item \textbf{Regional Harmonisation:}
The African Union should incorporate context-specific evaluation requirements into the implementation guidelines developed under Phase~I of the Continental AI Strategy~\cite{au2024aistrategy}, giving concrete content to its existing recommendation that member states review AI procurement standards. Without such specification, member states are left to develop evaluation requirements independently, reproducing the fragmentation already documented across the continent~\cite{njoroge2024potential,yilma2026strategy}. The evaluation framework proposed here, though developed for the Nigerian fintech context, may be structurally adaptable to other African markets with comparable digital banking and fintech ecosystems, including Kenya's mobile financial services sector and Ghana's digital banking and fintech landscape. This adaptability is not tested here and would require independent evaluation for each market's institutional and linguistic context. A shared continental evaluation methodology, adapted with local institutional content for each market, would reduce duplication of effort while preserving the context-specificity that generic global frameworks do not provide.

Comparable governance gaps have been documented in other African fintech markets. Kenya's fintech regulatory framework has similarly lagged behind its innovation pace, with formal oversight of emerging financial technologies typically arriving only after external pressure, not through proactive design~\cite{chambers2026kenya}. In South Africa, AI fraud detection systems developed for markets such as the United States face documented transferability barriers stemming from differences in fraud typologies, data availability, and regulatory frameworks, indicating that pre-deployment evaluation needs are market-specific even among comparably developed African economies~\cite{magangeni2026safraud}.

\item \textbf{Open Infrastructure as Governance Infrastructure:} Open evaluation toolkits function as governance infrastructure in their own right. They lower the cost of compliance for smaller fintech operators who cannot commission bespoke evaluations, and they extend the practical reach of any regulatory requirement adopted above. Where an evaluation dataset carries dual-use risk, as with the generation component of the SafeAlert framework demonstrated, gated access, granted only for a documented research or compliance purpose, allows regulators to maintain oversight without publicly distributing content that could enable fraud. A continental AI evaluation repository, maintained under the African Union or a designated technical body, would constitute a tractable near-term governance deliverable, consolidating evaluation toolkits and datasets developed across member states into a shared, independently auditable resource.

\end{enumerate}

\section{Conclusion}
\label{sec:conclusion}
This paper has examined why Nigerian and African continental AI
governance affirms safety in principle while leaving pre-deployment
evaluation unspecified. Using SafeAlert, we show the practical consequence of this gap: models that resist generic harmful content requests still produced complete fraud scripts under specific framing, and several misclassified most legitimate Nigerian bank communications as suspicious or fraudulent, a failure invisible to standard benchmarks and unaddressed by any existing regulatory requirement. Since this evaluation was conducted at minimal cost using a publicly available toolkit, the barrier is one of regulatory specification, not technical or financial constraint. We propose recommendations for the CBN, NITDA, SEC, and the AU as a starting point, with the underlying model of context-aware pre-deployment evaluation transferable to any African jurisdiction deploying AI in a domain-specific context.

\bibliographystyle{splncs04}
\bibliography{bibliography}

\end{document}